%% file: ms.tex
\documentclass{article}
\pdfoutput=1
\usepackage{amsmath}
\usepackage{graphicx}
\usepackage{amssymb}

\usepackage{xcolor}

\usepackage[breaklinks=true,bookmarks=false]{hyperref}

\usepackage{mathrsfs}
\usepackage{subfig}
\usepackage[export]{adjustbox}

\def\x{{\mathbf x}}
\def\eg{\emph{e.g.}} 
\def\ie{\emph{i.e.}}

\title{Leveraging Implicit Spatial Information in Global Features for Image Retrieval}

\author{Pierre Jacob$^1$,~
        David Picard$^{1,2}$,~
        Aymeric Histace$^1$,~
        Edouard Klein$^3$~\\
        $^1$ETIS UMR 8051, Universit{\'e} Paris Seine, UCP, ENSEA, CNRS, F-95000, Cergy, France \\
        $^2$Sorbonne Universit{\'e}s, Laboratoire d'Informatique de Paris 6, LIP6, F-75005 Paris, France \\
        $^3$C3N, P\^{o}le Judiciaire de la Gendarmerie Nationale, 5 boulevard de l'Hautil, 95000 Cergy, France\\
        \small \{pierre.jacob, picard, aymeric.histace\}@ensea.fr}

\date{}
\begin{document}

\maketitle
\begin{abstract}
    Most image retrieval methods use global features that aggregate local distinctive patterns into a single representation.
    However, the aggregation process destroys the relative spatial information by considering orderless sets of local descriptors.
    We propose to integrate relative spatial information into the aggregation process by taking into account co-occurrences of local patterns in a tensor framework.
    The resulting signature called Improved Spatial Tensor Aggregation (ISTA) is able to reach state of the art performances on well known datasets such as Holidays, Oxford5k and Paris6k.
    
\end{abstract}

\input{tex_files/introduction}

\input{tex_files/related_work}

\input{tex_files/method_overview}

\input{tex_files/experiments}

\input{tex_files/conclusion}

\bibliographystyle{unsrt}
\bibliography{tex_files/biblio}
\end{document}

%% file: tex_files/introduction.tex
\section{Introduction}
    In this paper, we are interested in image retrieval with a special focus on how to integrate spatial information in the process.
    In image retrieval, a collection of images is ranked by decreasing visual similarity with respect to a query, with the hope that relevant images are ranked first.
    This visual search engine is known as Content Based Image Retrieval (CBIR) and has been the subject of many improvements during the last decade.
    Applications of image retrieval includes copyright infringement detection~\cite{zhou2017effective}, geo-localization~\cite{Kim_2017_CVPR} or user interaction for shopping~\cite{whittlesearch_grauman_ijcv_2015} among others.

    Two competing strategies are popular for solving the image retrieval problem.
    The first one involves computing local descriptors in both the query and the target images and count the number of matching descriptors between the query and the target.
    A geometric consistency check is then applied to remove matches that are incoherent with the layout of both images (\textit{e.g.}, descriptors that are spatially close in the query should also be spatially close in the target)~\cite{jegou08}.
    As noted in \cite{li2015pairwise}, this spatial consistency check is crucial in descriptor matching techniques as it holds a major contribution to their good performance.
    However, despite efficient descriptors hashing techniques \cite{gong2013iterative,Jiang_2015_CVPR}, matching based strategies still come at a significant computational and storage cost.
    
    The second strategy addresses these shortcomings by computing a global representation for each image and then using a similarity measure on these representations as a proxy for the visual similarity \cite{jegou08_original_vlad, picard11_vlat, sanchez13_FV}.
    Recently, global features have become very competitive when compared to local descriptors matching on challenging datasets~\cite{arandjelovic_NetVLAD}.
    However, global features usually do not allow to perform a geometric consistency check because all spatial information is lost during the computation of the representation, and consequently do not benefit from the associated performance gain.
    
\begin{figure*}[th]
        \captionsetup[subfigure]{labelformat=empty}
        \begin{center}
            \subfloat{\includegraphics[width=0.2\linewidth]{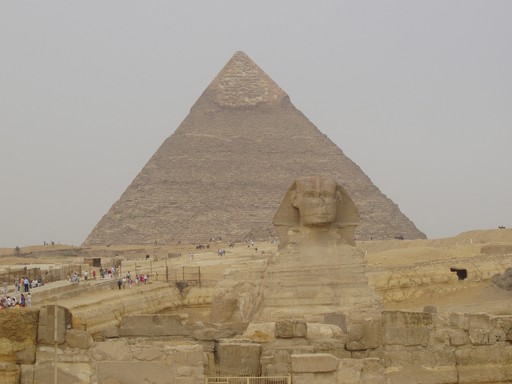}}
            \hfill
            \subfloat{\includegraphics[width=0.2\linewidth]{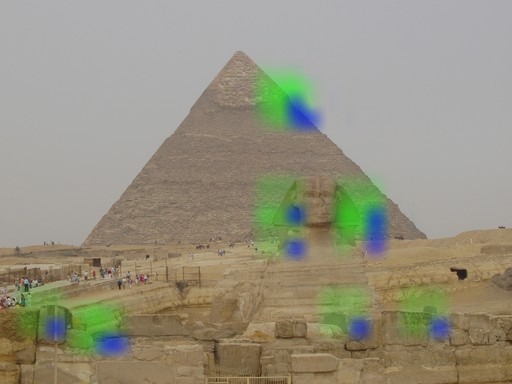}}
            \hfill
            \subfloat{\includegraphics[width=0.2\linewidth]{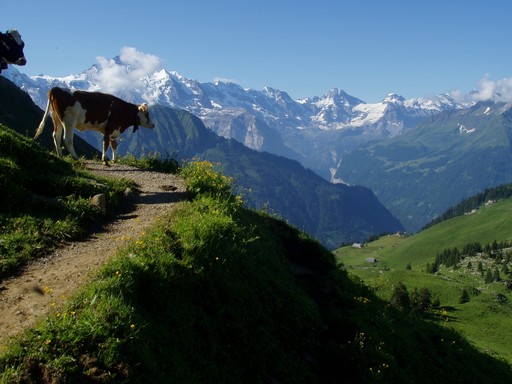}}
            \hfill
            \subfloat{\includegraphics[width=0.2\linewidth]{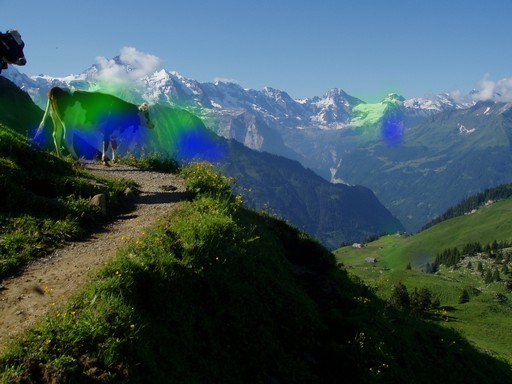}}
            \hfill \\
            
            \subfloat[original images]{\includegraphics[width=0.2\linewidth]{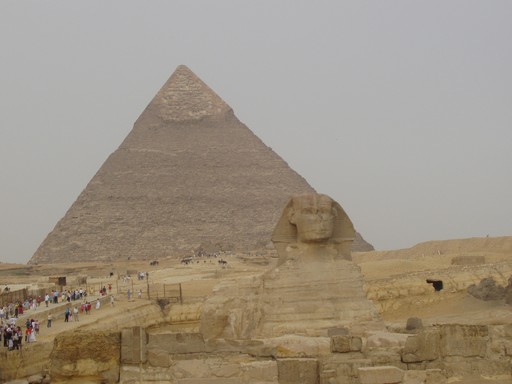}}
            \hfill
            \subfloat[1st best pair]{\includegraphics[width=0.2\linewidth]{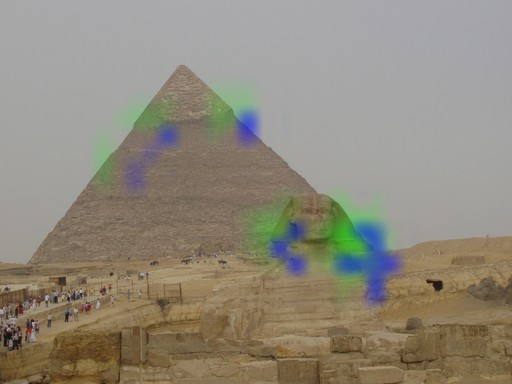}}
            \hfill
            \subfloat[original images]{\includegraphics[width=0.2\linewidth]{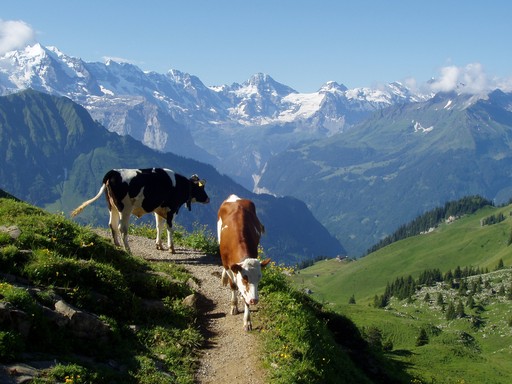}}
            \hfill
            \subfloat[1st best pair]{\includegraphics[width=0.2\linewidth]{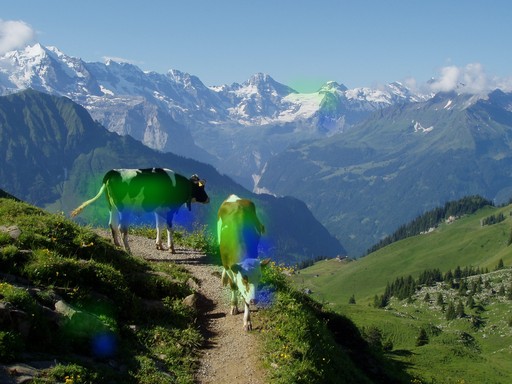}}
            \hfill \\

            \caption{Illustration of the implicit neighborhood matching. For both pairs of similar images, we show the pair of spatially coupled visual codebook entries that contributed the most to the similarity. We color in green the areas corresponding to descriptors belonging to the first codebook entry and in blue their respective neighbors belonging to the second codebook entry.This neighborhood encoding allows to focus on similar regions (\eg, the Sphinx head) with identical local spatial layout of the codebook entries. \emph{(Figure best viewed in color)}}
            \label{fig:illustration_2}
        \end{center}
    \end{figure*}

    In this paper, we intend to integrate spatial information into global features so as to perform implicit geometric consistency check when computing the similarity between the resulting features.
    We propose to integrate this information by modeling correlations between nearby features using a tensor framework, as illustrated by the results presented on \autoref{fig:illustration_2}. 
    Tensor embeddings have recently become popular for computing visual features \cite{do_faemb}.
    In particular, we build on the ideas proposed in the Spatial Tensor Aggregation (STA) of \cite{picard_STA} to propose a new global feature called ISTA.
    Our contributions are the following:
    \begin{itemize}
        \item[-] We correct all the flaws in the original STA, namely the lack of proper centering, alignment and normalization, leading to the Improved STA (ISTA).
        \item[-] We propose an adapted two-step dimension reduction method to cope with the high intermediate dimension of the STA.
        \item[-] Finally, our proposed model is able to obtain state of the art results on well known image retrieval datasets (Holidays, Oxford and Paris).
    \end{itemize}
        
    The remaining of this paper is as follows: First, we present the related work on global features computation for image retrieval. Then, we develop our Improved STA model for which we show all the corrections to the original STA as well as the proposed new steps and give intuition about their goals.
    Next, we present experiments comparing our model to the state of the art on three well known datasets, namely Holidays, Oxford5k and Paris6k.

%% file: tex_files/related_work.tex
\section{Related work}
    In content based image retrieval, aggregation algorithms are the current state of the art due to their very favorable trade off between computational complexity and performances.
    Efficient aggregation methods such as Fisher Vector~\cite{perronnin2010large,sanchez13_FV} and VLAD~\cite{jegou_compact_code} have been shown to provide very good results, especially when considering their improved version~\cite{arandjelovic_allAboutVLAD,delhumeau13_revisiting_vlad} combined with dimension reduction techniques~\cite{jegou_WKPCA,cao17_VSQ} and re-encoding methods~\cite{zepeda15_ESVM, rezende_SLEM}.
    
    Recently, Convolutional neural networks (CNN) have been shown to provide a strong baseline image representation for image retrieval even when the network is trained for a completely different task ~\cite{sharif2014cnn}.
    When used as features extractors \cite{gong2014multi}, CNN showed excellent improvements. 
    Finally, NetVLAD \cite{arandjelovic_NetVLAD} but also Deep VSQ \cite{cao17_VSQ} greatly improved the state-of-the-art by converting aggregation algorithms into differentiable neural network layers allowing to learn the representations in an end-to-end manner with dedicated loss functions.
    
    However, the main drawback of aggregation methods is that all spatial information is lost during the aggregation.
    Several methods were proposed in order to retrieve this spatial information either in absolute coordinates or in relative ones.
    Spatial Pyramid Matching (SPM) \cite{lazebnik06_spm} and MOP-CNN~\cite{gong2014multi} integrate spatial information in absolute coordinates by aggregating descriptors in cells of multiple scales following a recursive grid pattern.
    However, these methods see their dimension increased by a factor $4^l$ where $l$ is the number of scales, which becomes prohibitive.
    Since these methods model the entire image layout, there are not well suited for comparing images where the position of the regions of interest varies.
    
    Contrarily to these global methods, Spatial Tensor Aggregation (STA) \cite{picard_STA} is an aggregation method that integrates relative spatial information based on the linearization of the following matching kernel between spatially coupled pairs of descriptors:
    \begin{align}
        K(\mathfrak{B}_i, \mathfrak{B}_j) = \sum_{\substack{\x_{ri} \in \mathfrak{B}_i\\\x_{sj} \in \mathfrak{B}_j}}\sum_{\substack{\x_{u} \in \Omega(\x_{ri})\\\x_{v} \in \Omega(\x_{sj})}} k(\x_{ri},\x_{sj})k(\x_u, \x_v)
    \end{align}
    \vspace{-1pt}
    with $\mathfrak{B}_i = \{\x_{ri}\}$ being the set of local descriptors from image $i$ (resp. $\mathfrak{B}_j$ for image $j$), $\Omega(\x)$ being the set of descriptors in a spatial neighborhood centered on $\x$ and $k(\cdot, \cdot)$ a similarity measure between descriptors (\eg, based on a visual codebook).
    The embedding resulting from the linearization contains an implicit local geometric consistency check, in that the similarity is high only when both the descriptor and its neighbors are matched in the target image with preserved neighborhood properties.
    STA performs better than similar second-order information aggregations such as Fisher Vector \cite{sanchez13_FV} while offering an alternative to the spatial pyramid.
    While the STA idea is sound, the work of \cite{picard_STA} does not leverage local descriptors centering and processing which are known to have a significant impact~\cite{arandjelovic_allAboutVLAD,delhumeau13_revisiting_vlad}
    Furthermore, the tensor framework proposed in STA leads to signatures of too high a dimension to be of any practical use.
    The lack of proper normalization and dimension reduction therefore lead to non competitive results in the original paper. Because of non-trivial steps in the case of STA, we detail our propositions leading to a simple yet very competitive image feature in the next section.

%% file: tex_files/method_overview.tex
\section{Method overview}
    In this section, we detail our propositions to improve the STA aggregation method. In section \ref{centering} we propose an adapted descriptor processing. In section \ref{normalization} we investigate a proper normalization for the resulting feature. In section \ref{dim_reduc} we present a two step dimension reduction method.
    
    \subsection{Descriptors processing}\label{centering}
        In improved VLAD representations~\cite{delhumeau13_revisiting_vlad}, descriptors are processed using the following steps: First, the residual between the descriptor and its cluster center is computed. Then, this residual is projected into the eigenspace of the cluster (using PCA).
        
        In the case of the STA, we propose a similar approach. 
        Given a clustering $\left \{ \mathscr{C}_k\right \}_k$of the descriptors space, we first aggregate the residue between (i) a 4th order tensor computed from a descriptor and a given neighborhood ; (ii) the average of the 4th order tensors for the relevant cluster pair.
        More precisely, the 4th order aggregation tensor $\mathbf{T}(\mathfrak{B}_i)$ of the set of descriptors $\mathfrak{B}_i$ from image $i$ is computed with the following equation borrowed from \cite{picard_STA}:
        \begin{equation} \label{original_STA_model}
            \mathbf{T}(\mathfrak{B}_i) = \sum_{\substack{\mathbf{x}_{ri} \in \mathfrak{B}_i \\ \mathbf{x}_u \in \Omega(\mathbf{x}_{ri})}} \mathbf{h}(\mathbf{x}_{ri}) \otimes \mathbf{h}(\mathbf{x}_u) \otimes \mathbf{x}_{ri} \otimes \mathbf{x}_u
        \end{equation}
        where $\mathbf{h}(\cdot)$ is filled by 0 with a unique 1 at position $k$ if the descriptor belongs to the $k$-th cluster, $\Omega(\mathbf{x}_{ri})$ is the set of descriptors in a neighborhood of $\mathbf{x}_{ri}$, and $\otimes$ is the tensor (Kronecker) product.
        In other words, $\mathbf{T}(\mathfrak{B}_i)$ is the STA feature of image $i$ and computes the correlation matrices between pairs of nearby descriptors for all ordered pairs of clusters to which they belong.
        A more convenient form is to define the tensor $\mathbf{T}_{k, l}$ for a given pair of clusters $(k, l)$ using the following equation:
        \begin{equation}
            \mathbf{T}_{k, l}(\mathfrak{B}_i) = \sum_{\substack{\mathbf{x}_{ri} \in \mathfrak{B}_i \cap \mathscr{C}_k \\ \mathbf{x}_u \in \Omega(\mathbf{x}_{ri}) \cap \mathscr{C}_l}} \mathbf{x}_{ri} \mathbf{x}_u^T
        \end{equation}
        where $\mathscr{C}_j$ is the cluster $j$. By concatenating these tensors for all ordered pair of clusters, we recover the original 4th order tensor $\mathbf{T}$.
        
        To perform the centering, we compute the average 4-th order tensor $\mathfrak{T}$. With the same observation, we define $\mathfrak{T}_{k, l}$ the partial average tensor for the given pair of clusters $(k, l)$ and estimate it over a set of samples $\mathscr{S}_{k,l}$. $\mathscr{S}_{k,l}$ is the set of pairs $(\mathbf{x}_{r}, \mathbf{x}_{u})$ where $\mathbf{x}_{r}$ belongs to the $k$-th cluster and $\mathbf{x}_u$ belongs to the neighborhood of $\mathbf{x}_{r}$ and the $l$-th cluster:
        \begin{equation}
            \begin{aligned}
            \label{estimation_of_Tkl}
            \mathfrak{T}_{k, l} &= \frac{1}{|\mathscr{S}_{k,l}|}\sum_{\mathbf{x}_{r}, \mathbf{x}_u \in \mathscr{S}_{k,l} } \mathbf{x}_{r}\  \mathbf{x}_u^T
            \end{aligned}
        \end{equation}
        Finally, we are able to define the partial image representation $\mathbf{S}_{k, l}$ based on the aggregation of residues between the tensor $\mathbf{T}_{k,l}$ and its estimation $\mathfrak{T}_{k,l}$:
        \begin{equation}
            \label{sig_STA}
            \mathbf{S}_{k,l}(\mathfrak{B}_i) = \sum_{\substack{\mathbf{x}_{ri} \in \mathfrak{B}_i \cap \mathscr{C}_k \\ \mathbf{x}_u \in \Omega(\mathbf{x}_{ri}) \cap \mathscr{C}_l}}( \mathbf{x}_{ri} \  \mathbf{x}_u^T - \mathfrak{T}_{k, l})
        \end{equation}
        
        Since $\mathbf{T}_{k, l}(\mathfrak{B}_i)$ is not symmetric, projecting the descriptors into the eigenspace of their cluster pair is not as straightforward as for VLAD.
        We propose to use the Singular Values Decomposition (SVD) to perform this projection.
        For a given pair of clusters $(k, l)$, we compute the SVD of $\mathfrak{T}_{k, l}$:
        \begin{equation}
        \label{svd_Tkl}
            \mathfrak{T}_{k, l} = \mathbf{U}_{k, l} \mathbf{L}_{k, l} \mathbf{V}^T_{k, l} 
        \end{equation}
        
        We can inject Equation \eqref{svd_Tkl} into Equation \eqref{sig_STA} by multiplying by $\mathbf{U}^T$ on the left and by multiply by $\mathbf{V}$ on the right:
        \begin{equation}
            \label{svd_projected}
            \mathbf{S}^{(p)}_{k,l}(\mathfrak{B}_i) = \sum_{\substack{\mathbf{x}_{ri} \in \mathfrak{B}_i \cap \mathscr{C}_k \\ \mathbf{x}_u \in \Omega(\mathbf{x}_{ri}) \cap \mathscr{C}_l}} (\mathbf{U}_{k, l}^T\mathbf{x}_{ri}) \  (\mathbf{V}_{k, l}^T\mathbf{x}_u)^T - \mathbf{L}_{k, l}
        \end{equation}
        
        Using the result from the Equation \eqref{svd_projected} has two advantages:
        On the first hand, each descriptor is projected into the eigenspace which leads to decorrelated components in the resulting feature.
        On the second hand, only the singular values $\mathbf{L}$ are needed for the centering.
        Furthermore, we can keep only the largest singular values in order to remove irrelevant correlation between nearby descriptors while also reducing the size of the resulting features.
        In that aspect, we propose an adaptive strategy that keeps a variable number of components for each pair of clusters so as to keep a fixed amount of the explained variance (\eg, 80\%).
        The full signature is the concatenation of all $\mathbf{S}^{(p)}_{k,l}(\mathfrak{B}_i)$ for all pairs $(k,l)$.
    
        \begin{figure*}[ht]
    \captionsetup[subfigure]{labelformat=empty}
    \begin{center}
        \subfloat{\includegraphics[width=0.15\linewidth]{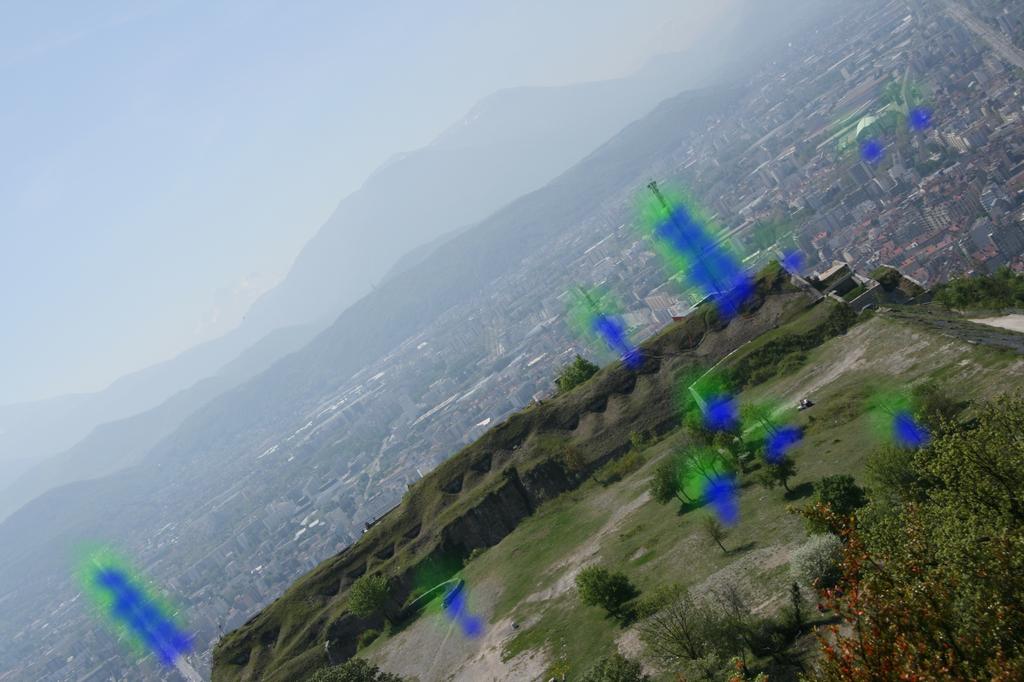}}
        \hfill
        \subfloat{\includegraphics[width=0.15\linewidth,cfbox=green 2pt 0pt 0pt]{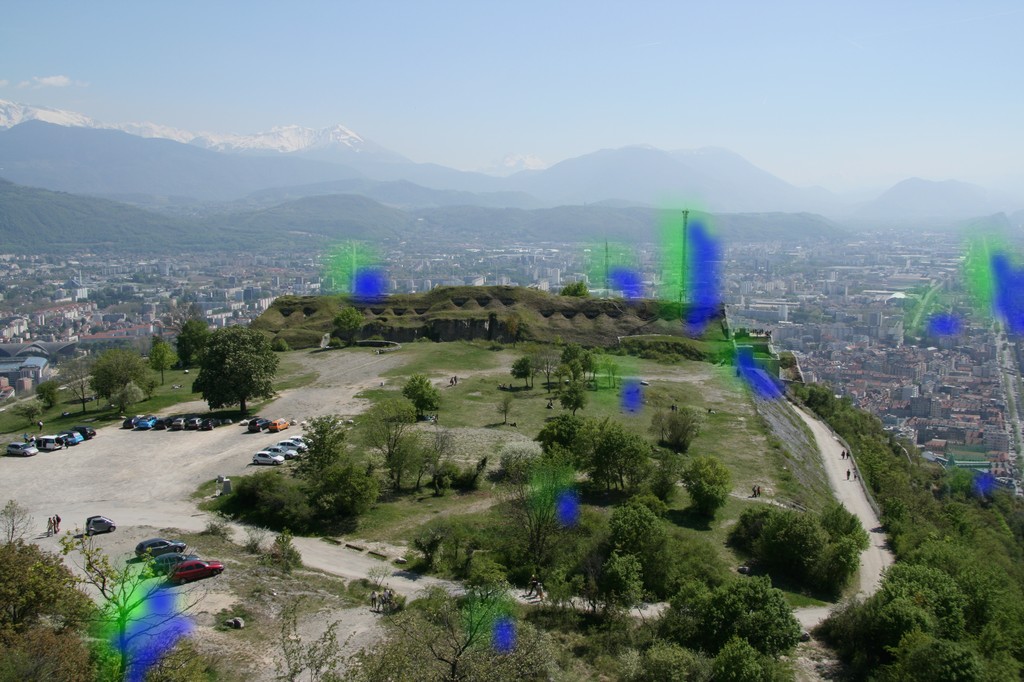}}
        \hfill
        \subfloat{\includegraphics[width=0.15\linewidth,cfbox=green 2pt 0pt 0pt]{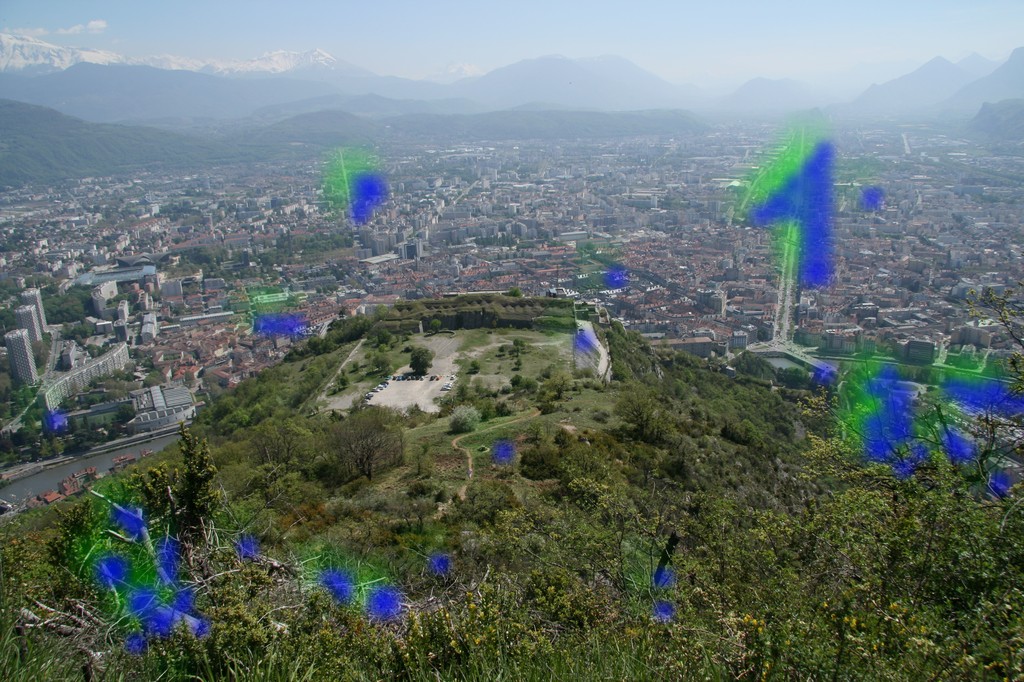}}
        \hfill
        \subfloat{\includegraphics[width=0.15\linewidth,cfbox=green 2pt 0pt 0pt]{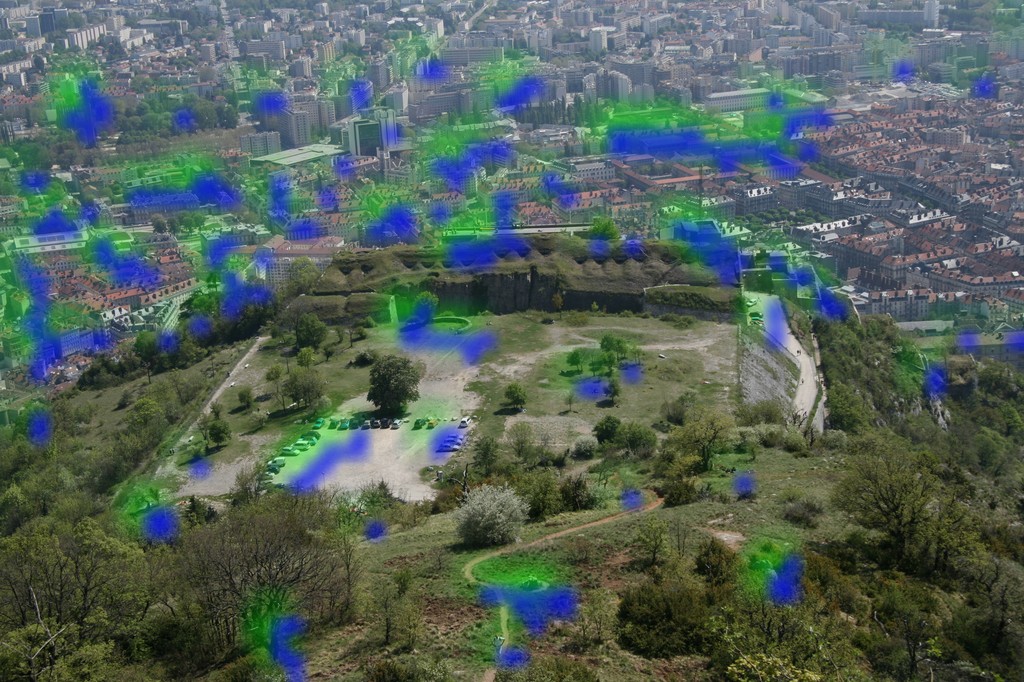}}
        \hfill
        \subfloat{\includegraphics[width=0.15\linewidth]{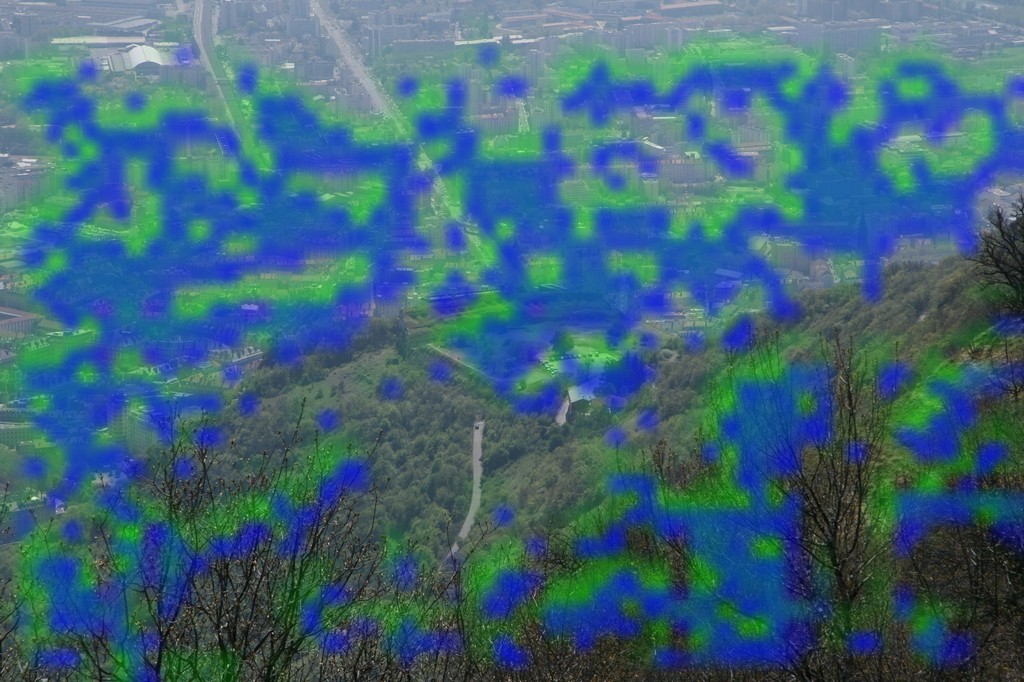}}
        \hfill
        \subfloat{\includegraphics[width=0.15\linewidth]{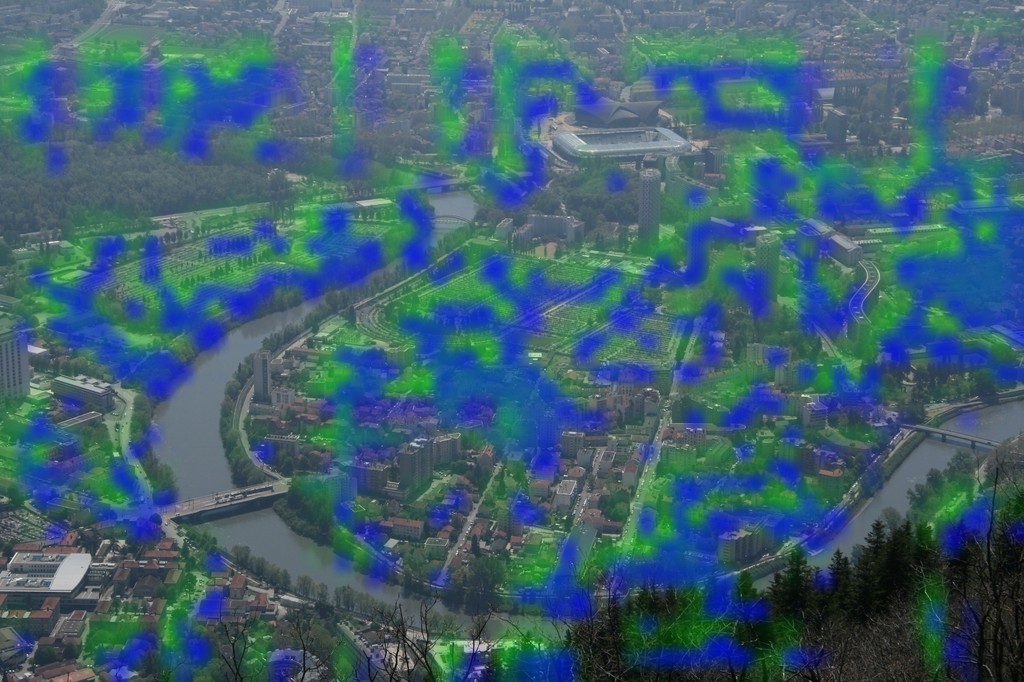}}
        \hfill \\
        
        \subfloat{\includegraphics[width=0.15\linewidth]{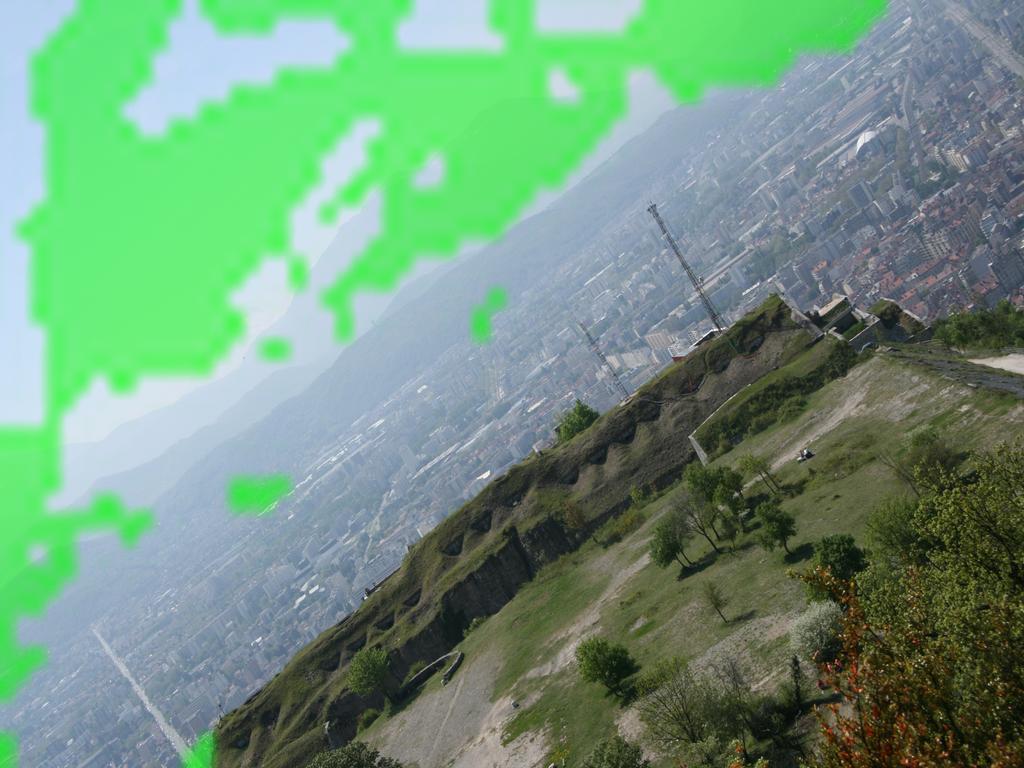}}
        \hfill
        \subfloat{\includegraphics[width=0.15\linewidth]{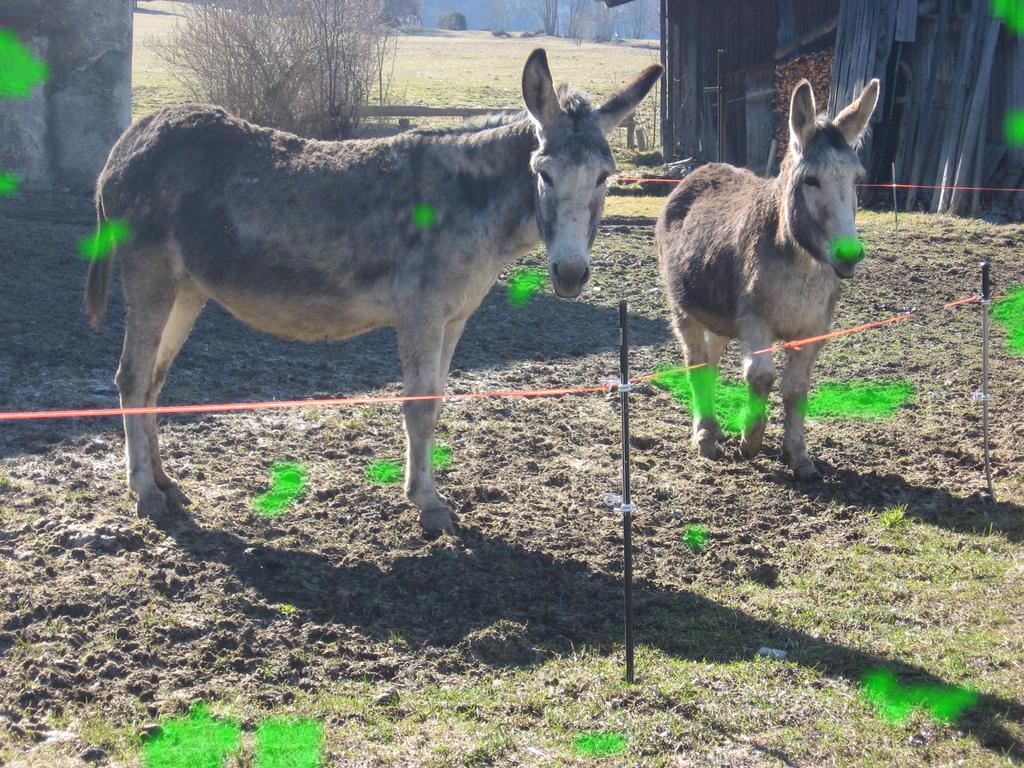}}
        \hfill
        \subfloat{\includegraphics[width=0.15\linewidth]{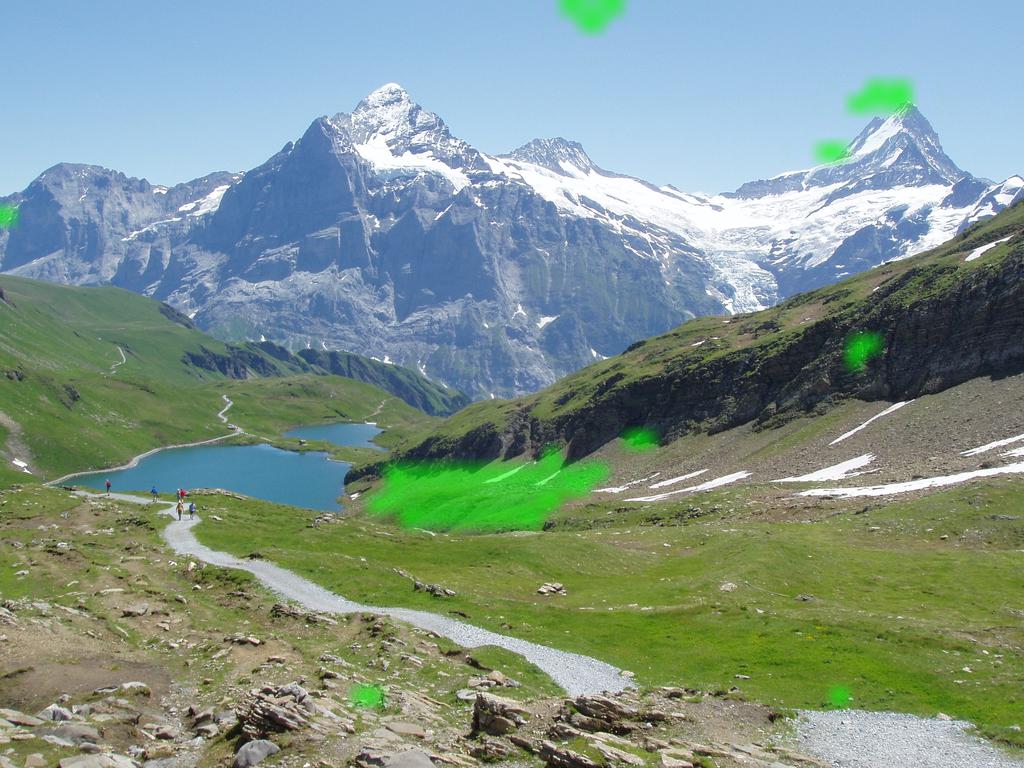}}
        \hfill
        \subfloat{\includegraphics[width=0.15\linewidth]{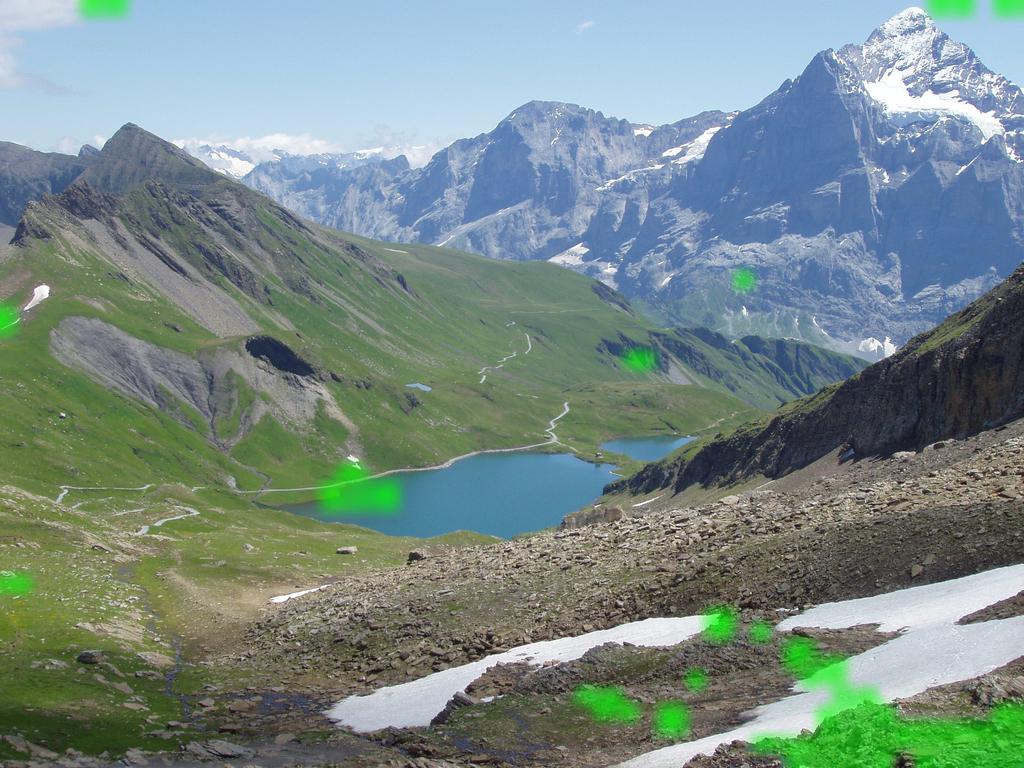}}
        \hfill
        \subfloat{\includegraphics[width=0.15\linewidth]{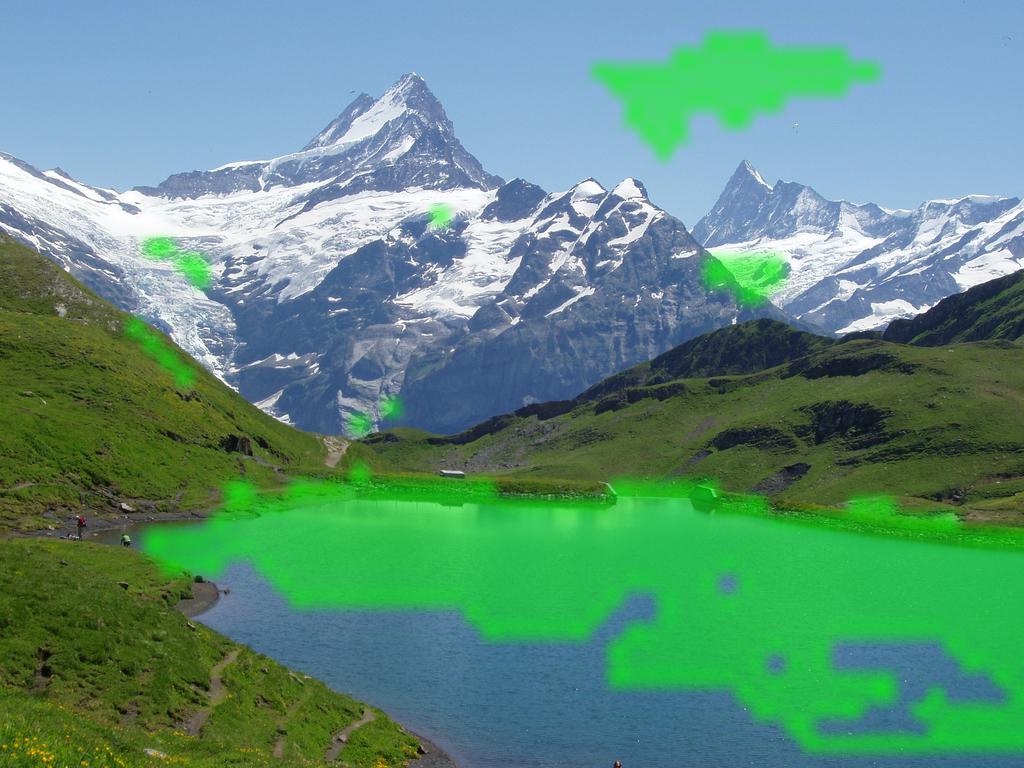}}
        \hfill
        \subfloat{\includegraphics[width=0.15\linewidth]{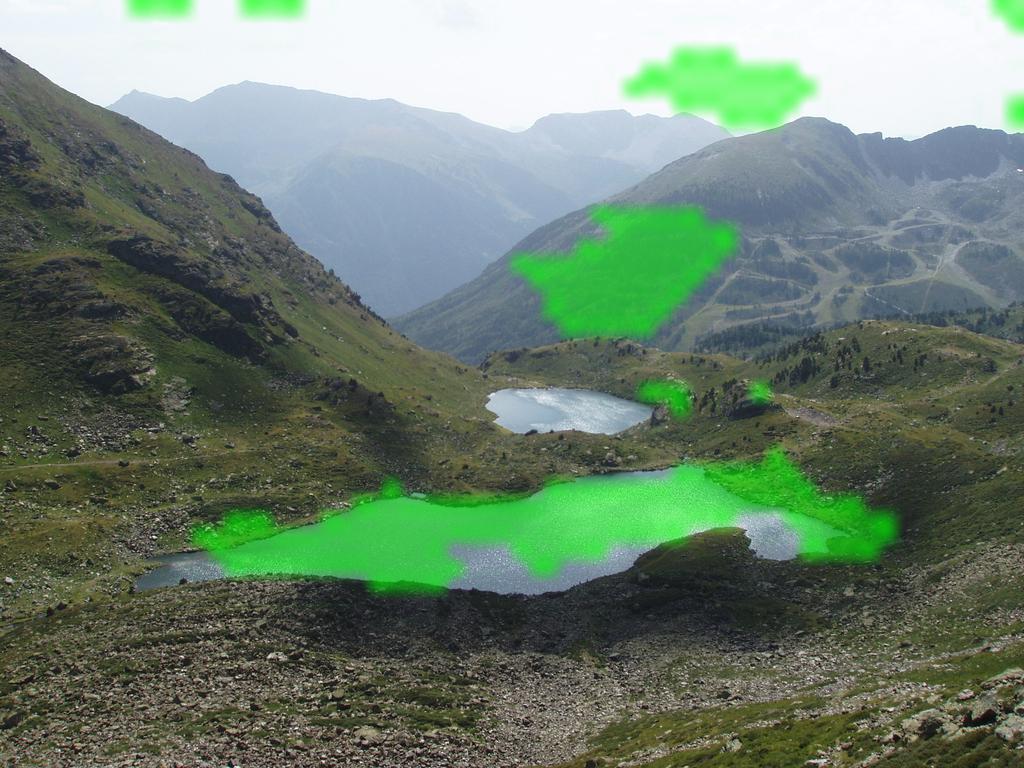}}
        \hfill \\

        \caption{Sample of queries improved by performing implicit spatial consistency check. Queries are on the first column, while the top 5 results retrieved by ISTA (first row) and NetVLAD (second row) respectively. Relevant images in the top 5 are outlined in green. Color masks in the images depict the region that contributed the most to the similarity. \emph{(Figure best viewed in color)}}
        \label{fig:illustration_gain}
    \end{center}
\end{figure*}

    \subsection{Normalization}\label{normalization}
        For a given number $N$ of clusters in the codebook, STA computes $N^2$ pair of clusters.
        Without a specific normalization, each tensor $\mathbf{T}_{k,l}$ has the same impact on the final representation.
        To reduce the influence of a potential noisy cluster, we propose to normalize separately the case $k=l$ from the case $k \neq l$.
        The reason behind the proposed approach is that both cases correspond to different contexts in the image.
        The case $k=l$ corresponds to self-similar regions in the image (\textit{e.g.}, textures), while the case $k\neq l$ corresponds to transitions between different patterns.
        As such, a cluster can have a negative impact in one context but not in the other.
        
        We propose to normalize in a cross-cluster way by considering similarly ranked eigen-components.
        Since descriptor pairs have been projected into the eigenspace of their cluster pair, processing components independently should not lead to any loss of information, with the added benefit of having a signature invariant to the number of descriptors and while keeping track of the distribution among cluster pairs.
    
        For a component $s_{k,l}(i,j)$ of $\mathbf{S}_{k,l}^{(p)}$, the normalization is computed as follow:
        
        \begin{equation}
            \label{normalization_eq}
            \forall i,j, s_{k,l}(i,j)=
            \begin{cases}
                \displaystyle\frac{s_{k,l}(i,j)}{\sqrt{\sum_n^{N} s_{n,n}^2(i,j)}} \text{ for } k=l \\
                \displaystyle\frac{s_{k,l}(i,j)}{\sqrt{\sum_{n \ne m}^{N, N} s_{n,m}^2(i,j)}}, \text{ else}
            \end{cases}
        \end{equation}

        This normalization is computed after a power normalization ($\alpha = 0.5$) and is followed by a $l_2$ normalization on the full image representation. 
        
    \subsection{Dimension reduction}\label{dim_reduc}
        After the centering and normalization, the resulting features have a dimension in the order of $ N^2D^2$ where $D$ is the number of component retained in the preprocessing. This usually leads to intermediate features of size close to $10^6$, which we now propose to reduce.
        Dimension reduction is usually achieved using PCA on a set of signatures by keeping components associated with the largest eigenvalues.
        In our case, PCA is not tractable neither with the classic approach nor with the Gram matrix decomposition due to the high dimension of the aggregated features.
        To cope with such high dimension, we propose a two step reduction scheme.
        First, we propose to perform a sparse reduction that considers components related to a specific cluster pair.
        We call this step Block Reduction due to the block diagonal structure of the resulting projection matrix.
        Then, we perform a full projection on the resulting vectors.
        
        The block reduction is performed independently for each part $\mathbf{S}_{k,l}^{(p)}$ of the normalized feature.
        This projection should map the feature block in a target space with fewer dimension while retaining the inner product.
        Indeed, since the inner product on the full signature is the sum of the inner products over all blocks $\mathbf{S}_{k,l}^{(p)}$, retaining the inner product on blocks is a sufficient condition to retain the full similarity.
        This is achieved by performing a low-rank approximation of a Gram matrix $\mathbf{G}_{k,l}$ computed on a large set of sampled blocks.
        
        In practice, we compute for each pair $(k,l)$ the projection matrix $\mathbf{P}_{k,l}$ using the SVD of a large set of blocks $\mathbf{S}_{k,l}^{(p)}$ and retaining a number of components proportional to the original number of dimension using the following equations:
        
        \begin{align*}
            \label{block_proj}
            \forall (i,j), \mathbf{G}_{k,l}(i,j)&= \left<\mathbf{S}_{k,l}(\mathfrak{B}_i) ; \mathbf{S}_{k,l}(\mathfrak{B}_j) \right> \\
            \mathbf{G}_{k,l}&=\mathbf{V}_{k,l}\mathbf{L}_{k,l}\mathbf{V}_{k,l}^T \\
            P_{k,l} &=\mathbf{L}^{\frac{1}{2}}_{k,l}\mathbf{V}_{k,l}^T(\mathbf{S}_{k,l}^{(p)})^T
        \end{align*}
        
        Each of these block-wise projection $\mathbf{P}_{k,l}$ is then zero padded to match the full size of the input features and the padded projections are concatenated.
        This result in a single sparse projection matrix $\mathbf{P} = \left[\mathbf{P}_{k,l}\right]_{k,l}$ with a block diagonal structure corresponding to the different pairs of clusters.
        
        Once the sparse projection is done, we perform the second reduction on the resulting features.
        Similarly to the first step, we aim at retaining the original inner product, which can be performed by finding a low-rank approximation of the Gram matrix using the SVD of a large set of features obtained by the sparse projection.
        
        As shown in \cite{jegou_WKPCA}, performing a whitening (\ie, equalizing the variance in the target space) at this stage can lead to a significant improvement, which is what we also observe.
        Remark that performing a whitening at this stage is only possible if no whitening was done during the block reduction stage as it would lead to all correlation between blocks being of equal importance.

%% file: tex_files/experiments.tex
\section{Experiments}\label{exp}
    In this section, we compare our proposed ISTA approach with recent approaches in the literature.
    We start by giving technical details and present the datasets on which the evaluation is performed before we comment on the results.
    
    \subsection{Experiments pipeline}
        As features extractor, we use the following off-the-shelf CNNs: VGG16 \cite{Simonyan14} (cut after block5) and MobileNet \cite{google17_mobilenet} (cut after block 11) both pre-trained on ImageNet \cite{imagenet}.
        All parameters are computed on 20k images taken from Places 365 \cite{zhou17_places2} validation set with a codebook of 32 visual words.
        In all our experiments, raw dimension is 670k, which corresponds to keeping 80\% of variance for VGG16 and 95\% for MobileNet in the preprocessing.
        The block projection are computed using 8192 images from Flickr100k and we kept 40\% of the original dimension.
        The full reduction is computed on 25k images from Flickr100k to reduce the final dimension to 22k.
        We compute the representations at 2 image resolutions (512px and 1024px) while conserving the image ratio and we sum the two representations.
        
        We test our model with 3 image retrieval datasets: INRIA Holidays \cite{jegou08} (1941 images, 500 queries), Oxford5k \cite{Philbin07_ox} (5062 images, 55 queries) and Paris6k \cite{Philbin08_paris} (6412 images, 55 queries).
        We report the mean average precision (mAP) for each datasets while using the full image as query. 
        
        \begin{table}[ht]
            \begin{center}
                \begin{tabular}{|c|c|c|c|} \hline
                    Method & Holidays & Oxford & Paris \\\hline\hline
                    Crow \cite{kalantidis2016cross} & 85.1 & 70.8  &  79.7 \\
                    NetVLAD \cite{arandjelovic_NetVLAD} & 88.3 & 69.1 & 78.5 \\ 
                    SLEM \cite{rezende_SLEM} & 91.7 & 71.7 & - \\\hline\hline
                    STA \cite{picard_STA} & 72.3  & - & -\\
                    ISTA VGG16 & 91.7 & 71.6 & 82.2\\
                    ISTA MobileNet & \textbf{94.4} & \textbf{77.1} & \textbf{88.8} \\ 
                    \hline
                \end{tabular}
            \end{center}
            \caption{Comparison with the state-of-the-art in mAP.}
            \label{table1_best_results}
        \end{table}
    
    \subsection{Results.}
        In this part, we compare ISTA on Holidays, Oxford5k and Paris6k against the state-of-the-art in Table \ref{table1_best_results}.
        Results with VGG16 as features extractor are similar to others methods that performs fine-tuning.
        However, these features were harder to compress than those extracted with MobileNet which leads to better results: +3\% mAP on Holidays and +5.5\% mAP on Oxford. We are able to outperform state-of-the-art networks on Holidays with 94.4\% of mAP versus 91.7\% mAP for NetVLAD with poly SLEM.
        On Oxford5k (resp. Paris6k), ISTA also outperforms the state-of-the-art obtained by the same methods by more than 5\% (resp. 10\%).
        
        Remark that most of the methods reported in \autoref{table1_best_results} perform a fine tuning over all parameters, whereas we do not perform it for ISTA.

        As an illustration, we show on \autoref{fig:illustration_gain} two queries that where among the most improved by ISTA over NetVLAD.
        For a given query, we show the associated top 5 ranked images for both methods, as well as masks of the regions that contributed the most to the ranking.
        As we can see, NetVLAD focuses on regions that may look similar taken independently from their context (like the sky in the first query), but that are not very distinctive.
        In contrast, our ISTA method focuses on strongly structured patterns that are much more distinctive.

%% file: tex_files/conclusion.tex
\section{Conclusion}

    In this paper, we propose the Improved Spatial Tensor Aggregation (ISTA) for aggregating local features into a single representation taylored for image retrieval.
    ISTA is based on a careful analysis of spatially coupled descriptors for which we provide essential centering, normalization and dimension reduction operations.
    All our contributions allow ISTA to obtain state of the art results on challenging datasets like Holidays, which show the soundness of the approach.
    
    In future work, ISTA can easily be adapted in a fully differentiable architecture like \cite{arandjelovic_NetVLAD}, which would allow the fine tuning of the full model.